\documentclass[letterpaper, 10 pt, conference]{ieeeconf}  

\IEEEoverridecommandlockouts                              

\usepackage{graphics} 
\usepackage{epsfig} 
\usepackage{mathptmx} 
\usepackage{times} 
\usepackage{amsmath} 
\usepackage{amssymb}  
\usepackage{pifont}
\usepackage{tcolorbox}
\usepackage{multirow}
\usepackage{booktabs}
\usepackage{caption}
\usepackage{graphicx}
\usepackage[pdftex,linkcolor=blue,citecolor=blue,backref=page]{hyperref}
\title{\LARGE \bf
User-Centric Object Navigation: A Benchmark with Integrated User Habits for Personalized Embodied Object Search
}

\author{Hongcheng Wang$^{*1}$, Jinyu Zhu$^{*2}$, Hao Dong$^{\ddag 1}$
\thanks{*These authors contributed equally.}
\thanks{$\ddag$Corresponding author.}
\thanks{$^{1}$Hongcheng Wang and Hao Dong are with School of Computer Science, Peking University, Beijing, China and PrimeBot {\tt\small \{whc.1999, hao.dong\}@pku.edu.cn}}
\thanks{$^{2}$Jinyu Zhu is with School of Electronics Engineering and Computer Science, Peking University, Beijing, China {\tt\small zhujinyu@stu.pku.edu.cn}}
\thanks{This work was supported by National Natural Science Foundation of China (62136001) and  The National Youth Talent Support Program (8200800081)}
}

\begin{document}

\maketitle
\thispagestyle{empty}
\pagestyle{empty}

\begin{abstract}
In the evolving field of robotics, the challenge of Object Navigation (ON) in household environments has attracted significant interest. Existing ON benchmarks typically place objects in locations guided by general scene priors, without accounting for the specific placement habits of individual users. This omission limits the adaptability of navigation agents in personalized household environments. 
To address this, we introduce User-centric Object Navigation (UcON), a new benchmark that incorporates user-specific object placement habits, referred to as user habits. This benchmark requires agents to leverage these user habits for more informed decision-making during navigation. UcON encompasses approximately 22,600 user habits across 489 object categories. UcON is, to our knowledge, the first benchmark that explicitly formalizes and evaluates habit-conditioned object navigation at scale and covers the widest range of target object categories.
Additionally, we propose a habit retrieval module to extract and utilize habits related to target objects, enabling agents to infer their likely locations more effectively. Experimental results demonstrate that current SOTA methods exhibit substantial performance degradation under habit-driven object placement, while integrating user habits consistently improves success rates. Code is available at \url{https://github.com/whcpumpkin/User-Centric-Object-Navigation}.

\end{abstract}


\IEEEpeerreviewmaketitle
\begin{figure*}[htbp]
    \centering
    \includegraphics[width=0.8\textwidth, trim=0 0 0 0, clip]{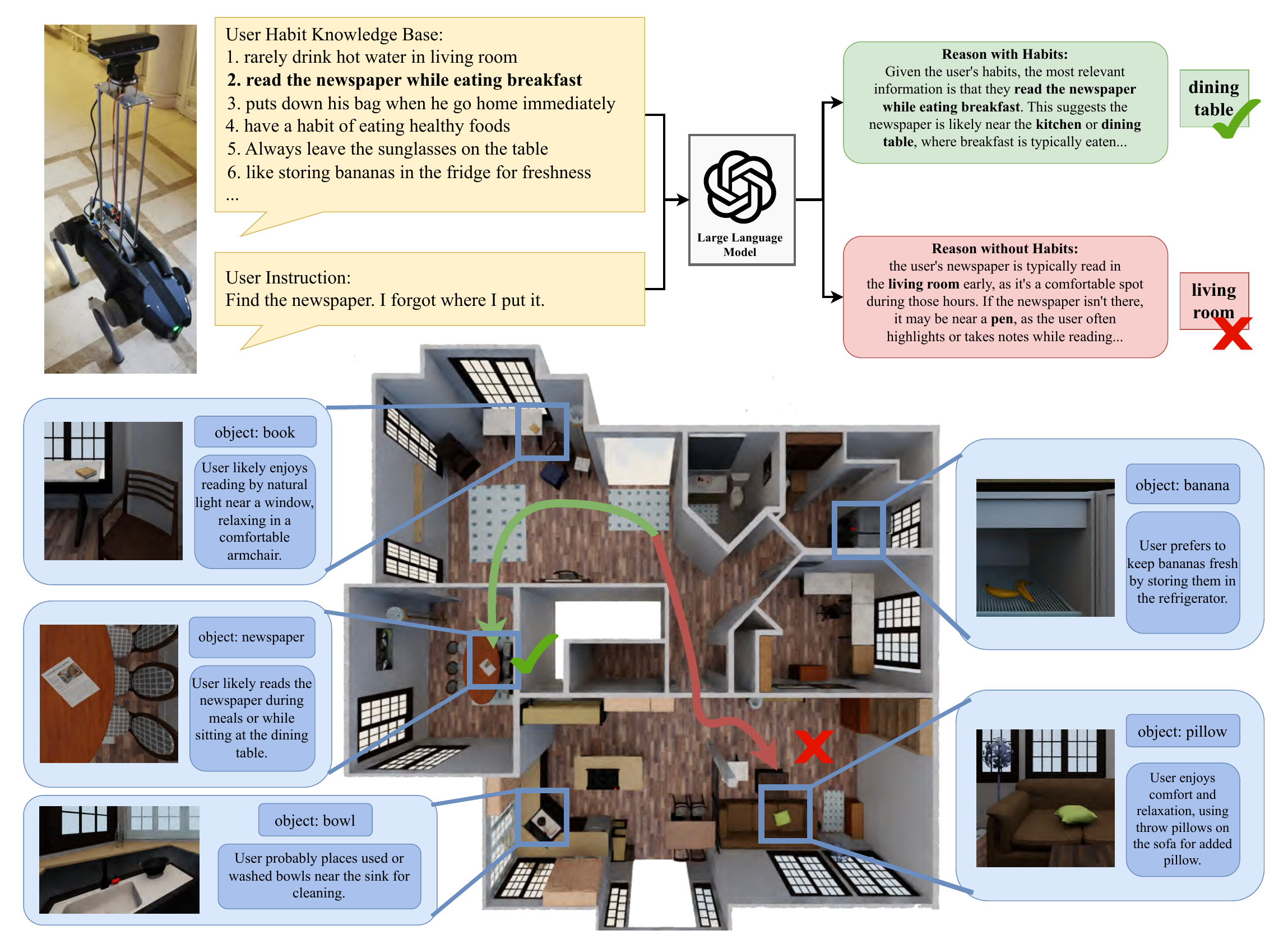} 
    \caption{\textbf{A Habit-shaped Scene and An Example of Reasoning.} In the bottom half of the figure we show five examples of objects being placed according to user habits (in practice the number will be higher). The top half is a presentation of the User Habit Knowledge Base and a comparison of LLM reasoning about whether or not to include habits.}
    \label{fig:teaser}
\end{figure*}
\section{Introduction}
``Where are my sunglasses? I remember that I wore it yesterday when eating lunch.'' Situations like this are very common in daily life: we may lose an object or simply forget its location. With the rapid development of robotics and embodied AI, many researchers focus on such an object search task (\emph{i.e.}, object navigation~\cite{cai2024bridging,chaplot2020object,du2021vtnet}).
Currently, some researchers~\cite{cai2024bridging,dorbala2023can, yinsg, long2024instructnav} leverage large language models (LLMs) such as GPT-4 for scene prior reasoning, and object detection models such as the YOLO series~\cite{redmon2016you} and SAM series~\cite{kirillov2023segment,ravi2024sam} for semantic map creation, with the aim of facilitating this task. Some researchers~\cite{yang2018visual,zhou2022improving,liang2021sscnav} focus on learning implicit object-scene association features to reason about the probability of occurrence of objects.
Most of them hold the assumption that the spatial distribution of objects adheres to common-sense norms; for instance, beds are typically found in bedrooms, while sofas are often arranged near TVs. This assumption is particularly rational for the mainstream object navigation benchmarks~\cite{yadav2022habitat,ramakrishnan2021hm3d,RoboTHOR} and our daily life in which the placement of target objects follows common-sense rules.

A natural question arises: do all objects in a household environment align with common sense? We cautiously propose that this may not always be the case. Consider, for instance, a scenario in which a person visits a friend’s house in search of a book. Intuitively, based on common sense, one might expect the book to be located in the study. However, the friend may have a habit of reading on the sofa and, consequently, might leave the book there after reading. In such a case, reliance on scene priors becomes insufficient for locating objects. This case naturally leads to another question: What strategies can aid in finding objects under such circumstances? We argue that \textbf{user habits} are the key to helping find objects in such cases.
In home environments, users constantly interact with objects within the scene. These interactions result in object placements that vary significantly among individuals. For example, one user might consistently return their glasses to a drawer after reading, while another might habitually leave them near the book on the sofa. Thus, we contend that the placement of objects is influenced not only by scene priors but also by \textbf{user habits}. To enable effective and long-term service, it is imperative for home service agents to not only recognize the habits of the household's residents, but also leverage these habits to facilitate efficient object search.

To systematically evaluate agents' capacity to exploit personalized behavioral patterns for object localization, we propose the User-centric Object Navigation Benchmark (UcON), a novel evaluation framework comprising 489 distinct object categories and over 22,600 human-object habits. Each episode initializes with a probabilistically sampled User Habit Knowledge Base (UHKB), containing object-habit associations that systematically modulate environmental configurations through habit-driven spatial constraints. As illustrated in Fig.~\ref{fig:teaser}, the lower panel demonstrates habit-conforming object placements, while the upper panel contrasts LLM-based reasoning paradigms for habit integration.
Detailed descriptions of the benchmark settings, including task generation, human test and realistic configurations, are provided in Sec~\ref{benchmark_setting}.
We argue that the core challenge of this task lies in effectively leveraging user habits. A critical question arises: is reasoning with user habits using large language models (LLMs) sufficient? Our experimental results indicate that relying solely on LLMs for reasoning is inadequate. Specifically, the experiments highlight several key challenges posed by UcON:
\begin{itemize} 
    \item Retrieving habits relevant to the target object from the User Habit Knowledge Base. 
    \item Filtering habits in the User Habit Knowledge Base that are inconsistent with the information obtained from the explored scene. 
    \item Leveraging user habits to guide efficient exploration of the environment. 
    \item Recognizing target objects across a diverse set of $489$ categories. 
\end{itemize}
To the best of our knowledge, UcON is the first comprehensive evaluation framework for habit-integrated navigation. It uniquely incorporates user habits, covers the widest range of object categories, and integrates the distinct challenges of habit retrieval, scene exploration, reasoning, and object recognition.

In this paper, we further propose a habit retrieval module to mitigate the challenges faced by LLMs in reasoning. This module simplifies the reasoning process by extracting task-relevant user habits based on the currently known scene information. Experimental results demonstrate that incorporating user habits improves the success rate of object navigation, while the proposed habit retrieval module further enhances navigation performance.

Our contributions are listed as follows:
\begin{itemize}
    \item We introduce the User-centric Object Navigation benchmark (UcON), featuring 489 object categories and $\sim$ 22,600 object-related habits. UcON surpasses existing benchmarks in scale and realism, aligning closely with real-world home service scenarios.
    \item We propose a simple yet effective Habit Retrieval Module (HRM) to enhance LLMs' ability to reason with user habits.
    \item We conduct extensive experiments using various SOTA object navigation methods. Results reveal the limitations of current approaches on the UcON benchmark while demonstrating that leveraging user habits improves performance, with further gains achieved by integrating the proposed HRM.
\end{itemize}

\begin{table}[]
\centering
\caption{A Comparison Between Benchmarks}
\begin{tabular}{l|c|c}
\hline
Benchmark         & {User Habits}  & \multicolumn{1}{l}{Object Category} \\ \hline\hline
Habitat ObjectNav~\cite{ramakrishnan2021hm3d}               &\ding{55}                     & 6               \\ \hline
MP3D ObjectNav~\cite{Matterport3D}                  &\ding{55}                     & 21              \\ \hline
RoboTHOR~\cite{RoboTHOR}                      & \ding{55}              & 43              \\ \hline
ProcTHOR~\cite{deitke2022️}                      & \ding{55}              & 108              \\ \hline
OVMM~\cite{homerobot}                            & \ding{55}             & 150               \\ \hline
UcON(Ours)             &\ding{51}                           & 489               \\ \hline
\hline
\end{tabular}
\end{table}
\section{Related Work}

\subsection{Object Navigation Benchmark}
Object Navigation (ON) is a core task in embodied AI, and its benchmarks are continually evolving. Prominent benchmarks like HM3D~\cite{ramakrishnan2021hm3d}, MP3D~\cite{Matterport3D}, and ProcTHOR~\cite{deitke2022️} have progressively increased the number of target object categories~\cite{habitatchallenge2023, homerobot}. However, these works primarily evaluate an agent's reasoning based on general scene priors—the conventional locations of objects within an environment. In contrast, our proposed UcON benchmark shifts the focus from scene priors to user habits, requiring the agent to learn and leverage personalized object placement tendencies to complete the navigation task.

\subsection{User-centric Embodied Task}

Incorporating user factors is a significant trend in the field of embodied AI. Recent work has begun to explore more user-centric tasks, where agents must adapt to human behaviors, instructions, and preferences to accomplish tasks such as social navigation~\cite{puig2023habitat}, collaborative rearrangement, or demand-driven navigation~\cite{ wang2024find,wang2024mo}.  BEHAVIOR-1K~\cite{li2024behavior} investigates 1000 tasks most relevant to user needs in home environments, offering insights into how agents can address practical household requirements.
Our work follows this direction, proposing that user habits are a critical and learnable signal that can significantly enhance an agent's navigation efficiency.

\subsection{Foundation Model in Object Navigation}

Foundation models~\cite{zhao2023survey,chowdhery2023palm,touvron2023llama2}, which have been trained on a large-scale dataset, show very good generalization and accuracy on the object navigation task. LGX~\cite{dorbala2023can} uses a large language model (LLM) to reason about the likely direction of the target object and then uses a visual language model for object detection. PixelNav~\cite{cai2024bridging} designs the prompt, then lets the LLM or VLM choose a direction and mark a pixel related to the targe object on the RGB input, then lets a low-level policy walk to the position described by the pixel. L3MVN~\cite{yu2023l3mvn} builds semantic maps and uses RoBERTa~\cite{liu2019robertarobustlyoptimizedbert} to determine which frontier has the highest probability of a target object. In our benchmarking, we try to use models that can run on consumer 24G graphics cards (e.g., 7B/13B LLMs/VLM), and our proposed habit retrieval model helps to improve the inference accuracy of small-sized LLMs.

\begin{figure}[htbp]
    \centering
    \includegraphics[width=\columnwidth, trim=00 0 130 60, clip]{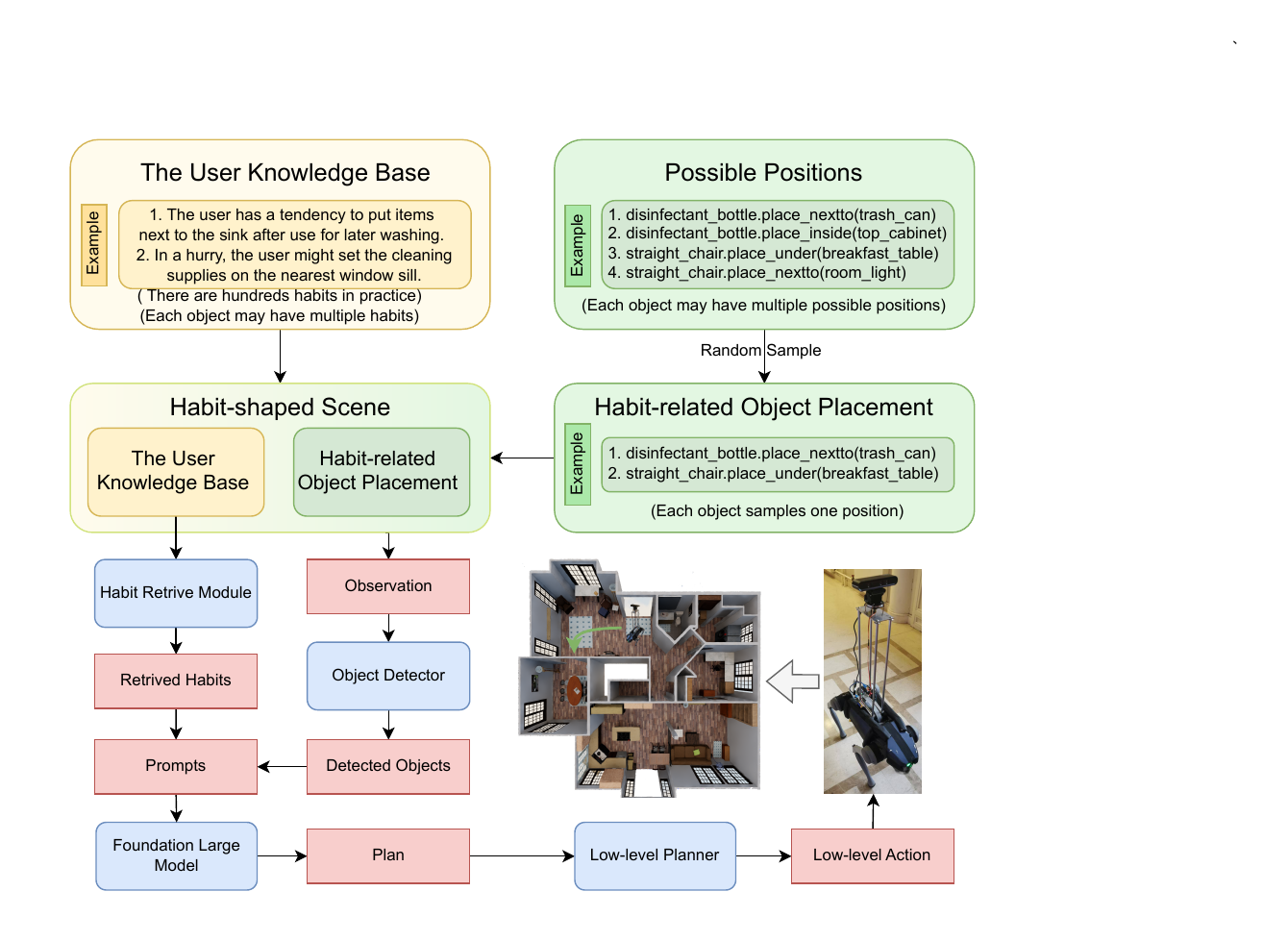} 
    \caption{\textbf{Task and Method Process Overview.} At the beginning, we sample $k$ objects and obtain their corresponding habits (each object may have multiply habits) to consist of the User Habit Knowledge Base, as well as all their possible positions (generated by GPT4, see Sec.~\ref{task_generation}). Each object is then sampled at a random position, and this position is used to modify the placement of the object to form a Habit-shaped Scene. A Habit Retrieval Module then retrieves the relevant habits, while an Object Detector identifies objects in the current observation. Information from both modules is then formatted by a prompter and fed into the Foundation Large Model to generate a plan.}
    \label{fig:TaskMethod}
\end{figure}

\section{Benchmark Setting}
\label{benchmark_setting}

\subsection{Task Statement}
\label{task_statement}

In each episode, an agent is randomly initialized within a \textbf{habit-shaped scene} (Sec.~\ref{simulator}). The agent's goal is to locate a target object from a category $o \in \mathbf{O}$, guided by a User Habit Knowledge Base, $\mathbf{H}=\{h_i\}_{i=1}^N$. This knowledge base is a large set ($N > 300$) of natural language habits, such as ``Users read books during meals.''

At each time step, the agent must select an action from the following set: $\mathrm{MoveAhead}$, $\mathrm{RotateRight}$, $\mathrm{RotateLeft}$, $\mathrm{LookUp}$, $\mathrm{LookDown}$, $\mathrm{Open}$, and $\mathrm{Done}$. The $\mathrm{Open}$ action opens all eligible containers within the agent's field of view and a $d_{open}=1\,\text{m}$ distance threshold. An episode, limited to 300 steps, is successful if the agent calls $\mathrm{Done}$ when the target object is visible and within a success threshold of $d_{succ}=1\,\text{m}$. Detailed descriptions of these actions are provided in Tab.~\ref{action_description}.

The core challenge is that $\mathbf{H}$ may contain contradictory habits from one or multiple users (e.g., coffee beans are stored near the kettle on weekdays but on the breakfast table on weekends). The agent must resolve such spatial conflicts by using evidence gathered during exploration to prioritize habits.

User-centric robotics involves two complementary challenges: habit acquisition (\emph{i.e.}, learning habits from long-term observation) and habit utilization (\emph{i.e.}, leveraging a known habit set for a task). This work intentionally focuses on habit utilization. We argue that effectively reasoning over and acting on a given habit knowledge base is a core capability that merits independent study; thus, UcON is the first benchmark designed to isolate and evaluate this downstream reasoning. Habit acquisition remains an important direction for future research.

\begin{table}[]
\centering

\caption{Detailed description of the effects of each action. }
\begin{tabular}{l|l}
\hline 
\multicolumn{1}{c|}{Action} & \multicolumn{1}{c}{Description}            \\ \hline\hline
$\mathrm{MoveAhead}$        & the agent moves forward 0.25 meters.       \\ \hline
$\mathrm{RotateRight}$      & the agent rotates 90 degrees to the right. \\ \hline
$\mathrm{RotateLeft}$       & the agent rotates 90 degrees to the left.  \\ \hline
$\mathrm{LookUp}$           & the agent's camera looks up 30 degrees.    \\ \hline
$\mathrm{LookDown}$         & the agent's camera looks down 30 degrees.  \\ \hline
$\mathrm{Open}$             & the agent open the closed container.       \\ \hline
$\mathrm{Done}$             & the episode ends.                          \\ \hline \hline                       
\end{tabular}
\label{action_description}
\end{table}

\subsection{Task Generation}
\label{task_generation}

We employ a three-step pipeline using the GPT-4 API to generate user habits and their corresponding object placements. 
First, we prompt GPT-4 with all object categories in a scene to generate natural language habits and associated object locations. Locations are expressed as one of four relative spatial relations: $\texttt{place\_nextto}$, $\texttt{place\_onTop}$, $\texttt{place\_inside}$, and $\texttt{place\_under}$. 
Second, the proposed placements are validated for physical plausibility within the simulation environment using a built-in API. For instance, a $\texttt{place\_onTop}$ relation is deemed invalid if an object lacks sufficient surface area to support another (\emph{e.g.}, placing a large box on top of a small vase).
Third, we use another prompt to have GPT-4 verify the semantic consistency between the generated habit text and the validated placement. This generation pipeline proves to be cost-effective, efficient, and scalable. See the Box~\ref{task_example} for a simple task example.

Note that in a specific episode, the User Habit Knowledge Base received by the agent not only contain habits related to the current target object but also other objects. Thus, the User Habit Knowledge Base contains a very large number of user habits (in the experiments there may be hundreds of habits), but only a very small fraction of them are relevant to the target object.

\begin{tcolorbox}[colframe=blue!75!black, colback=blue!10, title=Box III-B: An Example of Habits and Positions, fontupper=\small]
\label{task_example}

\textbf{Object}: coffee bean jar\\
\textbf{Habits}: \\
(1) Enjoys freshly brewed coffee every morning. \\
(2) Often forgets to return items to their storage place after use. \\
(3) Focuses on the perfect water temperature for coffee and might leave the jar by the kettle. \\
(4) Likes to keep coffee beans near the breakfast table for a quick start. \\
\textbf{Positions}: \\
coffee\_bean\_jar.place\_nextto(kettle)\\
coffee\_bean\_jar.place\_ontop(breakfast table)
\end{tcolorbox}

\subsection{Human Test}
We acknowledge that synthetically generated habits cannot fully capture the long-tail diversity and evolving nature of authentic human behavior. We adopt this pragmatic choice to establish UcON as a large-scale, reproducible, and controllable benchmark, which is difficult to achieve via real-world data collection at comparable scale.
To assess whether our synthetic habits and corresponding placements are \emph{plausible} to humans (as a sanity check on real-world relevance), we conducted a human evaluation study with 26 participants. Overall, 98.5\% of habits were judged as feasible, and 96.7\% of object placements were judged as consistent with the described habits. These results indicate that the generated instances are largely reasonable to humans, while we emphasize that this study evaluates plausibility rather than claiming to fully match the real-world distribution of human habits.

\subsection{Simulator}
\label{simulator}
We utilize the Omnigibson simulator~\cite{li2024behavior}, which provides 22 \textbf{initial scenes}. For each episode, we generate a \textbf{habit-shaped scene} by sampling $k$ objects and repositioning them according to associated user habits via the simulator's API. Object placement is performed sequentially to respect positional dependencies. To ensure environmental diversity, if multiple placements are specified by user habits, one is chosen at random for object instantiation, while all associated habits are added to the User Habit Knowledge Base, as referenced in Box~\ref{task_example}. Objects not present in the initial scenes are imported from the Omnigibson dataset. An example of this process is illustrated in Fig.~\ref{fig:TaskMethod}.

\subsection{Realistic Setting}
\label{realistic}
This benchmark targets object navigation in household environments, where user privacy is a primary concern. Reliance on external services introduces significant privacy risks and operational uncertainties, such as network latency and service instability. Consequently, we advocate for a self-contained, local solution to ensure data security and robust performance. To enhance practical feasibility, our approach is designed to operate efficiently on consumer-grade hardware (e.g., RTX 4090 or RTX 3090), addressing these critical privacy and stability challenges while minimizing computational overhead.

\section{Habit Retrieval Module}

While existing LLM-based Object Navigation (ON) methods can be adapted for User-Centric Object Navigation (UcON) by incorporating user habits into prompts, their reasoning performance degrades significantly with increased prompt length~\cite{xuretrieval,zhang2024infty,liu2024lost}. To mitigate this limitation, we introduce a Habit Retrieval Module inspired by Retrieval-Augmented Generation (RAG)~\cite{lewis2020retrieval}. This module selectively retrieves habits pertinent to a given target object from a User Habit Knowledge Base. Specifically, we use the BGE-M3 model~\cite{chen2024bge} to calculate the similarity between a query template (``Please retrieve the habit about \{TargetObject\}.'') and all stored habits, selecting the top-$p$ most relevant results for prompt integration. As depicted in Fig.~\ref{fig:TaskMethod}, we insert this module into the baseline process. Although the retrieval mechanism is simple, our primary contribution is demonstrating that augmenting prompts with relevant, retrieved habits enhances LLM-based navigation performance. Future work will focus on more sophisticated, context-aware retrieval methods that leverage scene-specific information to further improve efficiency and accuracy.

\section{Experiment}

\subsection{Simulation Experiments}
We conduct our simulation experiments on the Omnigibson simulator~\cite{li2024behavior}, which provides 22 scenes. At the beginning of each episode, we randomly select 20 objects, then import the habits corresponding to these objects into The User Habit Knowledge Base, and select 20 habits to modify the object locations. 

\subsection{Real-world Experiments}
Our real-world experiments were conducted in three environments, including small ($30m^2$) and large ($90m^2$) scenes, each with a manually designed User Habit Knowledge Base. The robotic platform (Fig.~\ref{fig:TaskMethod}) is a Unitree GO2 Edu equipped with an Astra Pro Plus RGB-D camera mounted on four 0.5m rods. To match the camera's Field of View (FoV), the robot's rotation angle was set to $60^\circ$. For computation, the GO2 connects via a Cat5 network cable to a ThinkPad E14 laptop, which in turn offloads processing via WiFi to an RTX 3090 workstation. We employed YOLO-worldv2-X as the object detector and a custom low-level path planner implemented with depth data and Breadth-First Search (BFS). PixelNav is used as the sole baseline for comparison.

\subsection{Baseline Methods}
We compare against several state-of-the-art (SOTA) methods. To reveal their limitations, we modify certain modules by substituting them with ground-truth (GT) inputs.

\begin{itemize}
    \item \textbf{PixelNav~\cite{cai2024bridging}} This method uses a multimodal LLM to select a navigation direction from multiple images. It has two variants: one inputs images directly, while the other inputs object labels extracted from the images. The agent then selects a target pixel in the chosen image for a low-level policy. In our implementation, we replace the object detectors (GroundingDINO and SAM) with GT object labels, the LLM (GPT-4) with LLaMA-3.2-11B-Vision-Instruct (variant 1) and LLaMA-3.1-8B-Instruct (variant 2), and the low-level policy with a GT path planner.

    \item \textbf{LGX~\cite{dorbala2023can}} LGX generates prompts from extracted object labels or scene descriptions, using an LLM's commonsense knowledge to select an obstacle-free direction. We replace its LLM (GPT-3) with GPT-4 and LLaMA-3.1-8B-Instruct, and its low-level policy with a GT path planner.

    \item \textbf{L3MVN~\cite{yu2023l3mvn}} This method predicts the likelihood of an object's occurrence using a RoBERTa model to guide navigation. We replace its local policy with a GT path planner.

    \item \textbf{VTN~\cite{du2021vtnet}} A closed-vocabulary, end-to-end navigation method that uses a Transformer-like model to fuse features and an LSTM network to predict actions.

    \item \textbf{ZSON~\cite{majumdar2022zson}} An open-vocabulary, end-to-end method that uses CLIP to encode the target object and a ResNet to encode observations, which are then fed to a policy network to predict actions.
\end{itemize}

\subsection{Metric}

\subsubsection{Success Rate (SR)} This task is considered successful when the agent executes $\mathrm{Done}$  and the target object appears in the field of view with a distance of less than 1 meter.

\subsubsection{Success weighted by Path Length (SPL)~\cite{anderson2018evaluation}} SPL is used to measure the efficiency of an agent in finding objects. The formula is as follows: 
\begin{equation}
    SPL=\frac{1}{N} \sum^N_{i=1} SR^{i} \frac{l_i}{max(p_i,l_i)}
\end{equation}
where $SR^{i}$ is the binary indictor of success in $i_{th}$ episode, $l_i$ is the length of the shortest path of the $i_{th}$ episode, and $p_i$ is the length of the path taken by the agent in the $i_{th}$ episode.

\begin{table*}[ht]
    \centering
    
    \caption{Performance Comparison Across Different Methods and Settings. Habit Level GT is the habit that is directly related to the target object which is not involved in the comparison because it is theoretically unavailable. Object Detector GT is also theoretically unavailable, but we still use it for experiments in order to test the upper bound of the methods. We report two metrics, SR and SPL, separated by ``/'', \emph{i.e}. ``SR/SPL''.}
    \begin{tabular}{lcc|cccc}
        \toprule
        \multirow{2}{*}{Method}   & \multirow{2}{*}{Language Model} & \multirow{2}{*}{Object Detector} & \multicolumn{4}{c}{Habit Level} \\
        \cmidrule{4-7}
                                  &                                     &                                  & None      & Full      & GT*      & Retrieval \\
        \midrule
        LGX                   & LLaMA-3.1-8B-Ins                    & GT                               & 20.2/14.8 & 25.2/18.4 & 20.4/14.6 & \textbf{28.2}/\textbf{21.3} \\
        LGX                   & LLaMA-3.1-8B-Ins                    & YOLO-Worldv2-X                       & 16.2/10.4 & 19.0/12.0 & 18.1/12.2 & \textbf{19.5}/\textbf{13.2} \\
        LGX                   & GPT-4                               & GT                               & 25.1/18.8 & 27.8/22.1 & 28.2/21.8 & \textbf{30.6}/\textbf{23.1} \\
        PixelNav              & LLaMA-3.2-11B-Vision                & GT                               & 15.6/13.0 & 17.3/\textbf{14.7} & 18.3/16.8 & \textbf{17.4}/14.6 \\
        PixelNav              & LLaMA-3.1-8B-Ins                    & GT                               & 16.9/14.9 & 17.1/14.3 & 17.2/15.1 & \textbf{20.5}/\textbf{17.1} \\
        L3MVN                 & RoBERTa-Large                       & GT                               & \textbf{10.6}/\textbf{9.8}  & /         & 10.0/9.0  & 10.3/9.3  \\
        VTN                   & /                                   & /                                & 2.1/1.4   & /         & /         & /         \\
        ZSON                  & /                                   & /                                & 1.0/0.7   & /         & /         & /         \\
        \bottomrule
    \end{tabular}
    \label{tab:performance_comparison}
\end{table*}

\subsection{Discussion}
In this section, we would like to discuss the following five questions.

\textbf{Q1: Can previous SOTA object method perform well in UcON \textbf{without user's habits} If they don't perform well, what limits their performance?}
Our experiments, detailed in Table~\ref{tab:performance_comparison}, reveal that all baseline methods exhibit a decline in success rate (SR) within the UcON benchmark compared to the SR in other ON benchmarks. End-to-end methods like VTN and ZSON are particularly ineffective. We attribute this failure to two main factors: 1) Their underlying assumption that object placement follows common-sense priors is violated in UcON, where locations are dictated by user habits. 2) These models are not architecturally designed to process and reason over complex, long-form natural language habits to infer object locations.

\begin{figure*}[h!]
    \centering
    \includegraphics[width=0.9\textwidth, trim=0 0 0 0, clip]{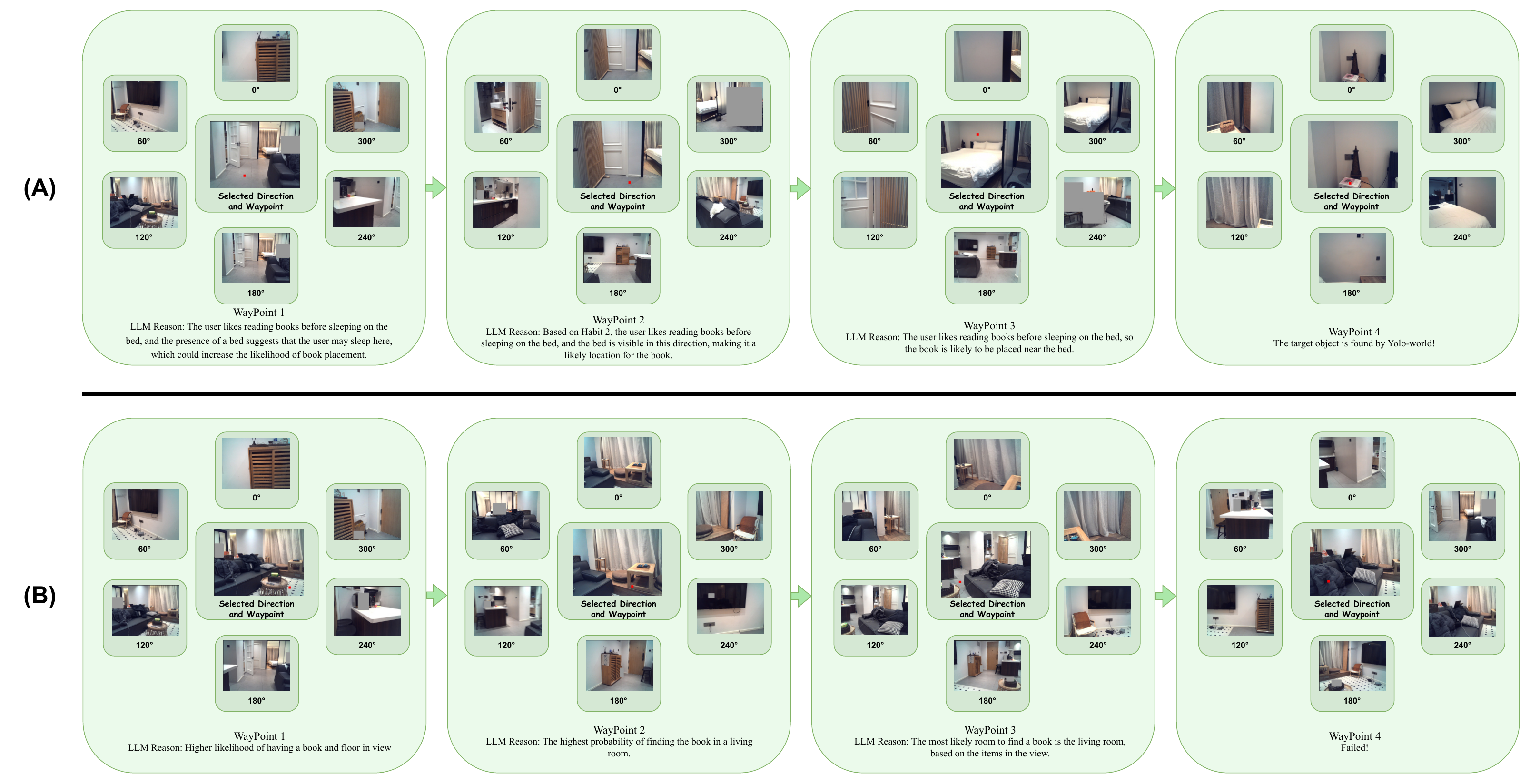}
    \caption{\textbf{A Case Study in The Large Environment} We show the waypoints and LLM reasoning results from one experiment. In this experiment, A book is initially invisible as a target object, but the agent is told about the user's habit of liking reading books before sleeping. The red squares represent the waypoints chosen by PixelNav. (A) The habit level is Retrieval. Although the books cannot be seen in the field of vision at first, based on the descriptions in the habit, the books are most likely placed in the bedroom, and then the direction in which the bed appears in the field of vision is most likely the bedroom. The book is also eventually found in the bedroom at the foot of the bed. (B) The habit level is None. Without user habits as a priori knowledge, LLM can only reason from common sense that books are more likely to be found in the living room compared to the kitchen. Eventually there is no book to be found after circling the living room a few times.}
    \label{fig:la_case}
\end{figure*}

Similarly, foundation model-based approaches (L3MVN, PixelNav, LGX) also suffer from performance degradation. This is due to a fundamental mismatch: the models' reasoning is grounded in common sense, whereas the environment's object placement is unconventional and habit-driven. This leads to inaccurate predictions for guiding exploration, as exemplified by the overly general reasoning observed in our real-world experiments (Fig.~\ref{fig:la_case}, Row B).

\textbf{Q2: Do user habits really help object navigation? Why do they help?}
The results for Full, GT, and Retrieval habit levels in Table~\ref{tab:performance_comparison} confirm that providing user habits substantially improves navigation success rates for most methods. The mechanism is straightforward: habits provide explicit textual clues about non-obvious object locations, which language models can leverage to guide the navigation policy toward more promising areas, thereby increasing exploration efficiency. We observe that more capable models, such as GPT-4, are more adept at utilizing these habits, leading to superior performance.

In contrast, L3MVN fails to benefit from habits, occasionally showing a slight performance decrease. We posit this is because its core RoBERTa-Large model, being an encoder-only architecture, lacks the strong inferential reasoning capabilities inherent in decoder-only models. Consequently, it cannot effectively process the provided habits. This limitation is further underscored by its inability to handle the context length of the Full Habit Level, for which its performance was not tested.

\textbf{Q3: Does our proposed habit retrieval module work? Why is it better than GT?} 
\label{Q3}
A counter-intuitive observation from our experiments is that the Retrieval habit setting frequently yields superior navigation performance compared to the GT setting. 
\emph{Here, the GT habit setting refers to the habit directly associated with the target object, rather than an oracle set of all task-optimal contextual information.}
We hypothesize that this phenomenon occurs because the retrieval mechanism surfaces habits that, while only indirectly related to the target object, provide crucial contextual information. 
This auxiliary knowledge allows the agent to reason more effectively about the spatial arrangement of other associated objects in the environment, leading to a more informed and efficient exploration strategy. 
We present a case study below to illustrate this hypothesis.

\begin{tcolorbox}[colframe=blue!75!black, colback=blue!10, title=Box V-E: A Case Study about Retrieval and GT,fontupper=\small] 
\label{GTR:case} 
\textbf{Object}: tea bag\\ \textbf{GT Habits}: \\ (1) While on phone calls and moving around the house, the tea bag could be left on a window sill or console table.\\ (2) While reading and sipping tea, the user might place the tea bag beside the armchair. \\ \textbf{Retrieved Habits}: \\ (1) While on phone calls and moving around the house, the tea bag could be left on a window sill or console table.\\ (2) While reading and sipping tea, the user might place the tea bag beside the armchair. \\ (3) The user keeps a teacup set on the breakfast table for regular use. \\ (4) I often read the newspaper at the breakfast table and then leave it on the near countertop. 
\end{tcolorbox}

As illustrated in the case study in Box~\ref{GTR:case}, the retrieved set includes habits that are indirectly related to the primary target, such as those concerning the ``breakfast table'' and ``countertop''. 
In this scenario, the LLM identifies these objects as relevant environmental landmarks, which in turn guides the agent to select the navigation direction containing them. 
This exemplifies a general principle: if a target object A is spatially correlated with a landmark object B, retrieving a habit pertaining to B can enable the agent to infer A's location by first navigating towards B.

\begin{figure*}[h!]
    \centering
    \includegraphics[width=0.9\textwidth, trim=0 0 0 0, clip]{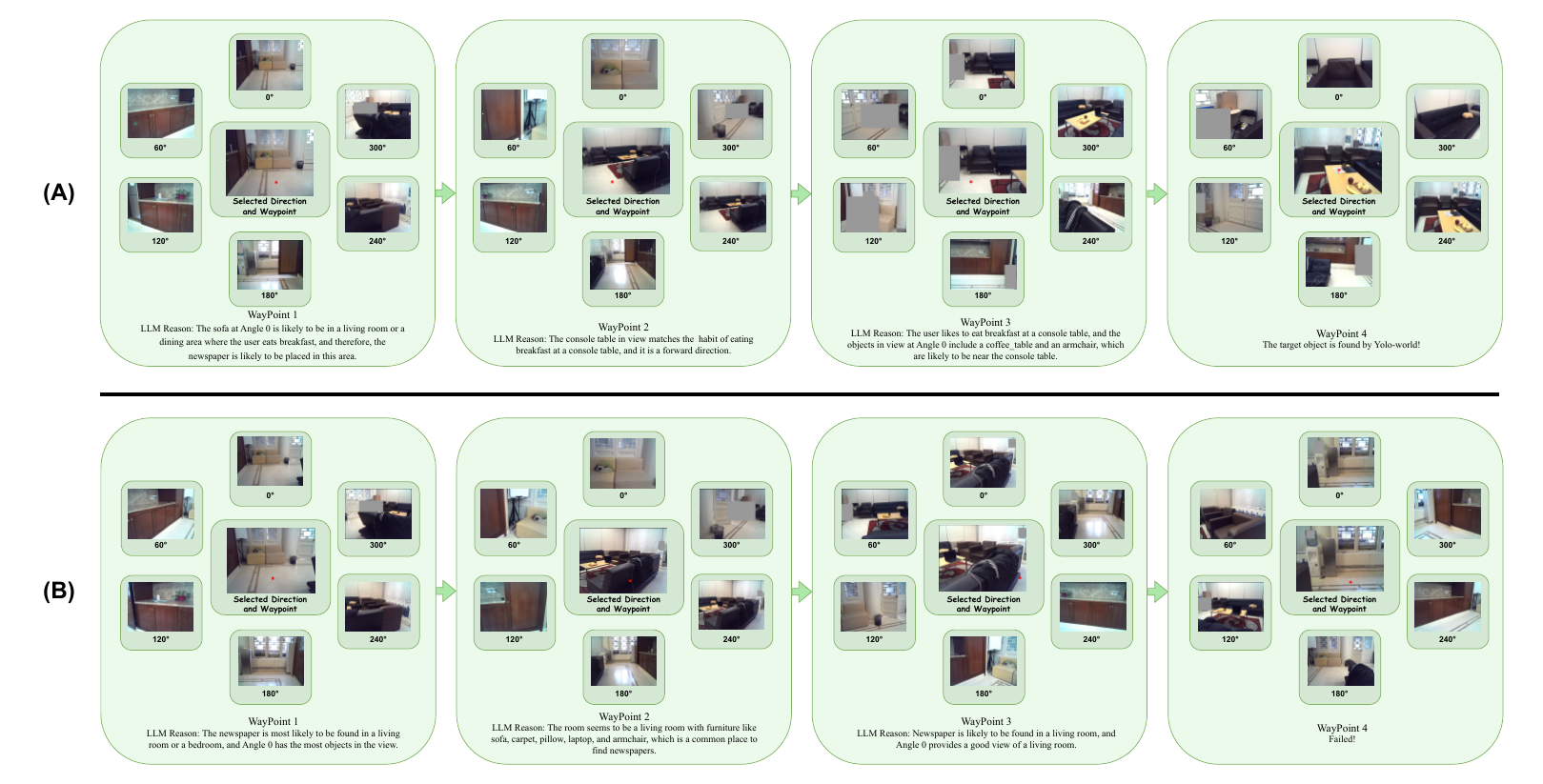}
    \caption{\textbf{A Case Study in The Small Environment} We show the waypoints and LLM reasoning results from one experiment. In this experiment, a newspaper, as a target object, is deliberately obscured by our laptops. The agent is told about the user's habit of enjoying breakfast on the console table while reading the newspaper. The red squares represent the waypoints chosen by PixelNav. 
    (A) The habit level is Retrieval. The agent first identifies the room where breakfast is served, then determines that the newspaper is nearby based on the console table in the field of view, and eventually finds the target. (B) The habit level is None. The agent can only use common sense to determine that the newspaper might be in the living room or bedroom, and then search aimlessly through the living room. In the end, the agent fails to find the target object}
    \label{fig:sma_case}
\end{figure*}

\textbf{Q4: Why is it that with the GT object detector provided, and the habits retrieved, the success rate of agents finding objects is still not high?} The relatively modest success rates, even with GT components, are themselves a meaningful finding. They highlight the intrinsic difficulty of the proposed user-centric reasoning task and expose fundamental limitations of current state-of-the-art LLMs.
In particular, the diversity of user habits (Box III-B) often yields multiple plausible locations for a single object, creating substantial ambiguity. Resolving this ambiguity requires capabilities that current models still lack: (1) multi-step planning to prioritize and search likely locations efficiently; (2) persistent exploration memory to avoid redundant searches; and (3) evidence-based belief updating to prune hypotheses and habits based on exploration feedback.
Consequently, UcON functions not only as an evaluation platform for existing methods, but also as a challenging benchmark that highlights key open problems and motivates future research in robotic reasoning.

\textbf{Q5: Do user habits still work for navigation in real-world experiments?} We do some case studies on two habit level, None and Retrieval. Shown in Fig.~\ref{fig:la_case} and Fig.~\ref{fig:sma_case}, we find that user habits can still help navigation in real scenarios. The a priori knowledge provided by the user's habits will directly guide the agent's behavior when the object placement is not quite common sense.






\section{Limitation}
Our work is a first step toward user-centric object navigation and has several limitations. First, to build a large-scale, reproducible benchmark, we rely on synthetically generated habits; while we validate plausibility and semantic consistency, synthetic habits may miss long-tail diversity and temporal drift in real households. Second, we focus on \emph{habit utilization} given a habit memory, leaving continual habit acquisition and updating from long-term interaction to future work. Finally, our Habit Retrieval Module is a proof-of-concept and does not explicitly model rich context (e.g., time/location) or uncertainty under partial observability; as suggested by our analysis (Q3), robust performance will likely require context-conditioned reasoning, multi-step planning, and evidence-based belief updating / active search.

\section{Conclusion}
\label{sec:conclusion}
In this paper, we proposed UcON, a novel benchmark with 489 target objects that integrates user habits as \textit{a priori} knowledge to aid navigation, particularly for objects not placed according to common sense. The key challenge in UcON is efficiently leveraging these habits. To this end, we also introduced a Habit Retrieval Module (HRM) to retrieve target-relevant habits from a knowledge base. Experiments confirm that user habits enhance navigation success, and our proposed HRM yields further significant improvements. What we hope this research will bring to the community is how to better utilize user habits and how to better serve humans.









\bibliographystyle{IEEEtran}
\bibliography{references}

\end{document}